\documentclass{article}

\usepackage[english]{babel}
 
\usepackage[utf8x]{inputenc}
\usepackage[T1]{fontenc}
\usepackage{natbib}
\usepackage{subcaption}
\usepackage{booktabs}
\usepackage{multirow}
\usepackage{url}
\usepackage{soul}
\usepackage[dvipsnames]{xcolor}
\usepackage{tikz}
\usepackage{fontawesome5}
\usepackage{tikz}
 \usetikzlibrary{patterns,patterns.meta,shadows,shadows.blur,arrows.meta}

\usetikzlibrary{calc,quotes,decorations.pathreplacing,decorations.text}

\usepackage{paralist}
\usepackage[stable]{footmisc}
\usepackage{adjustbox}

\usepackage{ifthen}
\newcommand{\CC}[1][]{$\text{C\hspace{-.25ex}}^{_{_{_{++}}}}
\ifthenelse{\equal{#1}{}}{}{\text{\hspace{-.625ex}#1}}$}

\usepackage{bm}
\usepackage{bbm}
\usepackage{amsmath}
\usepackage[all,warning]{onlyamsmath}
\usepackage{amssymb}
\usepackage{amsthm}
\usepackage{amsfonts}
\usepackage{thmtools}	
\usepackage[euro]{isonums}

\usepackage[mathic=true]{mathtools}
\usepackage{fixmath}
\usepackage{siunitx}
\usepackage{dirtytalk}

\usepackage{mleftright}
\usepackage{stmaryrd}
\usepackage{xfrac}

\usepackage{thm-restate}

\let\originalleft\mleft
\let\originalright\mright
\renewcommand{\mleft}{\mathopen{}\mathclose\bgroup\originalleft}
\renewcommand{\mright}{\aftergroup\egroup\originalright}

\usepackage{pifont}

\usepackage{enumitem}
\setlist[enumerate]{itemsep=0.1ex, topsep=0.1\topsep}
\setlist[description]{itemsep=0.1ex, topsep=0.1\topsep}
\setlist[itemize]{itemsep=0.1ex, topsep=0.1\topsep}

\makeatletter
\def\thmt@refnamewithcomma #1#2#3,#4,#5\@nil{%
\@xa\def\csname\thmt@envname #1utorefname\endcsname{#3}%
\ifcsname #2refname\endcsname
\csname #2refname\expandafter\endcsname\expandafter{\thmt@envname}{#3}{#4}%
\fi
}
\makeatother

\usepackage[pagebackref,
pdfa,
hidelinks,
pdftex, 
pdfdisplaydoctitle,
pdfpagelabels,
pdfauthor={Christopher Morris},
pdftitle={},
pdfsubject={},
pdfkeywords={MPNNs, GNNs, benchmarks, dataset, foundation model},
pdfproducer={Latex with the hyperref package},
pdfcreator={pdflatex}
]{hyperref}

\usepackage{todonotes}
\usepackage[capitalise,noabbrev]{cleveref}

\theoremstyle{definition}

\theoremstyle{remark}

\usepackage[activate={true,nocompatibility},final,kerning=true]{microtype}
\usepackage{ellipsis}

\usepackage[accepted]{icml_arxiv}

\newcommand{\new}[1]{\emph{#1}}

\icmltitlerunning{Graph Learning Will Lose Relevance Due To Poor Benchmarks}

\begin{document}

\twocolumn[
\icmltitle{Position: Graph Learning Will Lose Relevance Due To Poor Benchmarks}

\icmlsetsymbol{equal}{*}

\begin{icmlauthorlist}
\icmlauthor{Maya Bechler-Speicher}{equal,tau,meta}
\icmlauthor{Ben Finkelshtein}{equal,ox}
\icmlauthor{Fabrizio Frasca}{equal,tech}
\icmlauthor{Luis Müller}{equal,ac}
\icmlauthor{Jan Tönshoff}{equal,ac}
\icmlauthor{Antoine Siraudin}{ac}
\icmlauthor{Viktor Zaverkin}{nec}
\icmlauthor{Michael M. Bronstein}{ox}
\icmlauthor{Mathias Niepert}{st}
\icmlauthor{Bryan Perozzi}{go}
\icmlauthor{Mikhail Galkin}{go}
\icmlauthor{Christopher Morris}{ac}
\end{icmlauthorlist}

\icmlaffiliation{tau}{Tel-Aviv University}
\icmlaffiliation{meta}{Meta}
\icmlaffiliation{tech}{Technion - Israel Institute of Technology}
\icmlaffiliation{ox}{University of Oxford}
\icmlaffiliation{ac}{RWTH Aachen University}
\icmlaffiliation{nec}{NEC Laboratories Europe}
\icmlaffiliation{st}{University of Stuttgart}
\icmlaffiliation{go}{Google Research}

\icmlcorrespondingauthor{Maya Bachler-Speicher}{mayab4@mail.tau.ac.il}
\icmlcorrespondingauthor{Luis Müller}{luis.mueller@cs.rwth-aachen.de}

\icmlkeywords{graph neural networks, GNNs, MPNN, benchmarks, datasets}

\vskip 0.3in
]

\printAffiliationsAndNotice{\icmlEqualContribution} 

\begin{abstract}
While machine learning on graphs has demonstrated promise in drug design and molecular property prediction, significant benchmarking challenges hinder its further progress and relevance. Current benchmarking practices often lack focus on transformative, real-world applications, favoring narrow domains like two-dimensional molecular graphs over broader, impactful areas such as combinatorial optimization, relational databases, or chip design. Additionally, many benchmark datasets poorly represent the underlying data, leading to inadequate abstractions and misaligned use cases. Fragmented evaluations and an excessive focus on accuracy further exacerbate these issues, incentivizing overfitting rather than fostering generalizable insights. These limitations have prevented the development of truly useful graph foundation models. \emph{This position paper calls for a paradigm shift toward more meaningful benchmarks, rigorous evaluation protocols, and stronger collaboration with domain experts to drive impactful and reliable advances in graph learning research, unlocking the potential of graph learning.}
\end{abstract}

\section{Introduction}

Graphs are versatile mathematical structures capable of modeling complex interactions among entities across a wide range of disciplines, including the life sciences~\citep{Won+2023}, social sciences~\citep{Eas+2010}, and optimization~\citep{Cap+2021}, underlining the need for specialized machine-learning methods to extract meaningful insights from graph-structured data. Hence, in recent years, \emph{message-passing graph neural networks} (MPNNs)~\citep{Gil+2017} have emerged as the leading architecture for machine learning on graphs. These architectures---and, more broadly, \new{graph neural networks} (GNNs)---have become prominent topics at top-tier machine learning conferences,\footnote{\url{http://tinyurl.com/mpn89vju}} demonstrating promising performance across a diverse range of applications. Notable examples include their role in breakthroughs such as discovering new antibiotics~\citep{Sto+2020, Won+2023} and advancements in weather forecasting~\citep{Lam+2023}.

Despite these successes, we contend that for graph learning to remain \emph{relevant} and \emph{impactful}, current benchmarks need to be aligned with such truly \emph{transformative real-world applications}. While various benchmarks have been proposed, many existing datasets focus on narrow domains or address problems with questionable practical relevance. For instance, popular benchmarks frequently feature two-dimensional molecular graphs~\citep{hu2020ogb,Mor+2020}, neglecting critical three-dimensional geometric structures. Additionally, many studies report state-of-the-art results on (synthetic) datasets like \textsc{Zinc}~\citep{Dwi+2022}, which lack sufficient (real-world) justification for their graph-based approach, further complicating their utility. Empirical studies in graph learning often suffer from methodological shortcomings. Inconsistent dataset splits and evaluation protocols across studies undermine the validity of comparisons, while the reliance on small datasets frequently results in high-variance outcomes with limited statistical significance. Due to these limitations and the scarcity of sufficiently large and diverse datasets, MPNNs and GNNs have shown limited evidence of scalability to large pre-trained or foundation models.

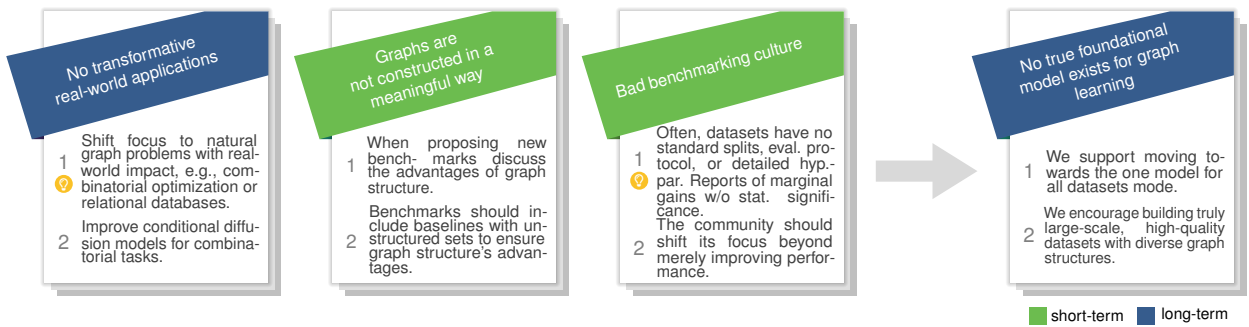
\begin{figure*}[t]
	\begin{center}

\begin{tikzpicture}[transform shape, scale=0.85]
\clip (-0.7,0.15) rectangle (19,-5.3);

\definecolor{c6}{HTML}{47337E}
\definecolor{c5}{HTML}{365C8D}
\definecolor{c4}{HTML}{277F8E}
\definecolor{c3}{HTML}{1FA187}
\definecolor{c2}{HTML}{4AC16D}
\definecolor{c1}{HTML}{6CBC4F}
\definecolor{yllw}{HTML}{FAC127}

%1
\begin{scope}[shift={(0,0)}]
\draw[fill=c6!70!black,draw=none] (-0.2,-2) -- (0.1,-2) -- (0.1,-1.8) -- cycle;
\draw[draw=none, fill=gray!50!white] (0.2,-0.2) rectangle (3.7,-4.45);
\draw[draw=none, fill=gray!25!white] (0.1,-0.1) rectangle (3.6,-4.35);
\draw[draw=none, fill=white,blur shadow={shadow scale = 1, shadow blur steps=30, shadow xshift=0pt, shadow yshift=0pt}] (0,0) rectangle (3.5,-4.25);

\draw[fill=c5,draw=none] (3.5,-1) -- (3.5,0) -- (2.5,0) -- (-0.6,-0.7) -- (-0.2,-2) -- cycle;

\node[align=center, anchor=center,white,rotate=15] at (1.4,-0.9) {\scalebox{0.75}{\parbox{4cm}{\centering \textsf{No transformative} \\ \textsf{real-world applications}}}};

\node[align=center,gray!50!black] at (1.95,-2.5) {\scalebox{0.5}{\parbox{5.5cm}{\textsf{\Large Shift focus to natural graph problems with real-world impact, e.g., combinatorial optimization or relational databases.
}}}};
\node[gray!100!black] at (0.3,-2.35) {\scalebox{0.9}{\textsf{1}}};
\node[gray!100!black] at (0.3,-3.6) {\scalebox{0.9}{\textsf{2}}};

\node[align=center,gray!50!black] at (1.95,-3.6) {\scalebox{0.5}{\parbox{5.5cm}{\textsf{\Large Improve conditional diffusion models for combinatorial tasks.
}}}};

\draw[draw=none,fill=yllw, rounded corners = 14pt] (0.3,-2.7) circle [radius=4pt];
\node[align=center, anchor=center,white] at (0.3,-2.7) {\scalebox{0.55}{\faLightbulb[regular]}};

\end{scope}

%2
\begin{scope}[shift={(4.5,0)}]
\draw[fill=c3!70!black,draw=none] (-0.2,-2) -- (0.1,-2) -- (0.1,-1.8) -- cycle;
\draw[draw=none, fill=gray!50!white] (0.2,-0.2) rectangle (3.7,-4.45);
\draw[draw=none, fill=gray!25!white] (0.1,-0.1) rectangle (3.6,-4.35);
\draw[draw=none, fill=white,blur shadow={shadow scale = 1, shadow blur steps=30, shadow xshift=0pt, shadow yshift=0pt}] (0,0) rectangle (3.5,-4.25);

\draw[fill=c1,draw=none] (3.5,-1) -- (3.5,0) -- (2.5,0) -- (-0.6,-0.7) -- (-0.2,-2) -- cycle;

\node[align=center, anchor=center,white,rotate=15] at (1.4,-0.9) {\scalebox{0.75}{\parbox{4cm}{\centering \textsf{Graphs are} \\ \textsf{not constructed
in a} \\ \textsf{meaningful way}}}};

\node[align=center,gray!50!black] at (1.95,-2.4) {\scalebox{0.49}{\parbox{5.7cm}{\textsf{\Large When proposing new bench- marks discuss the advantages of graph structure.}}}};
\node[gray!100!black] at (0.3,-2.4) {\scalebox{0.85}{\textsf{1}}};
\node[gray!100!black] at (0.3,-3.6) {\scalebox{0.85}{\textsf{2}}};

\node[align=center,gray!50!black] at (1.95,-3.6) {\scalebox{0.5}{\parbox{5.5cm}{\textsf{\Large Benchmarks should include baselines with unstructured sets to ensure graph structure's advantages.
}}}};

\end{scope}

%3
\begin{scope}[shift={(9,0)}]
\draw[fill=c1!70!black,draw=none] (-0.2,-2) -- (0.1,-2) -- (0.1,-1.8) -- cycle;
\draw[draw=none, fill=gray!50!white] (0.2,-0.2) rectangle (3.7,-4.45);
\draw[draw=none, fill=gray!25!white] (0.1,-0.1) rectangle (3.6,-4.35);
\draw[draw=none, fill=white,blur shadow={shadow scale = 1, shadow blur steps=30, shadow xshift=0pt, shadow yshift=0pt}] (0,0) rectangle (3.5,-4.25);

\draw[fill=c1,draw=none] (3.5,-1) -- (3.5,0) -- (2.5,0) -- (-0.6,-0.7) -- (-0.2,-2) -- cycle;

\node[align=center, anchor=center,white,rotate=15] at (1.4,-0.9) {\scalebox{0.75}{\parbox{4cm}{\centering \textsf{Bad benchmarking culture}}}};

\node[align=center,gray!50!black] at (1.95,-2.5) 
{\scalebox{0.5}{\parbox{5.5cm}{\textsf{\Large \!\! Often, datasets have no standard splits, eval. protocol, or detailed hyp.-par. Reports of marginal gains w/o stat. significance.}}}};
\node[gray!100!black] at (0.3,-2.3) {\scalebox{0.9}{\textsf{1}}};
\node[gray!100!black] at (0.3,-3.7) {\scalebox{0.9}{\textsf{2}}};

\node[align=center,gray!50!black] at (1.95,-3.7) {\scalebox{0.5}{\parbox{5.5cm}{\textsf{\Large The community should shift its focus beyond merely improving performance.}}}};

\draw[draw=none,fill=yllw, rounded corners = 14pt] (0.3,-2.65) circle [radius=4pt];
\node[align=center, anchor=center,white] at (0.3,-2.65) {\scalebox{0.55}{\faLightbulb[regular]}};

\end{scope}

\begin{scope}[shift={(13,-2.5)},scale=0.8]
\draw[gray!30,line width=8pt,-{Triangle[angle=60:7mm]}] (0,0) -- (2,0);
\end{scope}

%4
\begin{scope}[shift={(15.1,0)}]
\draw[fill=c3!70!black,draw=none] (-0.2,-2) -- (0.1,-2) -- (0.1,-1.8) -- cycle;
\draw[draw=none, fill=gray!50!white] (0.2,-0.2) rectangle (3.7,-4.45);
\draw[draw=none, fill=gray!25!white] (0.1,-0.1) rectangle (3.6,-4.35);
\draw[draw=none, fill=white,blur shadow={shadow scale = 1, shadow blur steps=30, shadow xshift=0pt, shadow yshift=0pt}] (0,0) rectangle (3.5,-4.25);

\draw[fill=c5,draw=none] (3.5,-1) -- (3.5,0) -- (2.5,0) -- (-0.6,-0.7) -- (-0.2,-2) -- cycle;

\node[align=center, anchor=center,white,rotate=15] at (1.4,-0.9) {\scalebox{0.75}{\parbox{4cm}{\centering \textsf{No true foundational model
exists for graph learning}}}};

\node[align=center,gray!50!black] at (1.95,-2.5) {\scalebox{0.5}{\parbox{5.5cm}{\textsf{\Large We support moving towards
the one model for all datasets mode.}}}};
\node[gray!100!black] at (0.3,-2.5) {\scalebox{0.9}{\textsf{1}}};
\node[gray!100!black] at (0.3,-3.5) {\scalebox{0.9}{\textsf{2}}};

\node[align=center,gray!50!black] at (1.95,-3.5) {\scalebox{0.47}{\parbox{6cm}{\textsf{\Large We encourage building truly
large-scale, high-quality datasets with diverse graph structures.}}}};

\end{scope}

\begin{scope}[shift={(23.8,-6.8)}]

\pgfdeclarehorizontalshading{viridisgradient}{100bp}{%
  color(0bp)=(c1);
  color(20bp)=(c1);
  color(40bp)=(c3);
  color(60bp)=(c4);
  color(80bp)=(c5);
  color(100bp)=(c6)
}
\scalebox{0.7}{\fill[draw=none, fill=c1] (-1.8,0.2) rectangle (-1.4,-0.2);

\fill[draw=none, fill=c5] (0.6,0.2) rectangle (1,-0.2);
\node[black] at (-0.5,0) {\scalebox{1}{\textsf{short-term}}};
\node[black] at (1.9,0) {\scalebox{1}{\textsf{long-term}}};}

\end{scope}
    
\end{tikzpicture}
\vspace{-16pt}
	\end{center}
    \caption{\label{fig:overview} Overview of the current challenges in benchmarking for graph learning and possible remedies.}
\end{figure*}

\paragraph{Present work} In this position paper, we argue that graph learning must significantly revise its current datasets and benchmarking practices to remain impactful and relevant; see~\cref{fig:overview} for an overview. Specifically, we
\begin{enumerate}
\item discuss the current shortcomings in graph learning benchmarks, including the lack of transformative real-world problems, an overfocus on specific data modalities, and fragmented evaluation protocols, resulting in the absence of true foundation models for graph data;
\item propose possible remedies to address these shortcomings, offering actionable recommendations for the graph learning community; and
\item based on our assessment of current graph benchmarks, we tune a variety of new baselines and reference models on molecular prediction tasks, large-scale heterophilic datasets, and study in- and cross-domain transfer in a pre-training/fine-tuning setup.  \end{enumerate}

\emph{\textbf{Overall, in this position paper, we argue that the benchmarking aspect of graph learning requires a significant revision for the field to stay impactful and relevant, including the design of current datasets, the investigated data modalities, and current benchmarking practices.}} 

In the remaining part of this section, we provide a critical overview of the current state of the field. In the following four sections, we highlight four current shortcomings of graph datasets and benchmarking practices and their possible remedies.

\paragraph{Basic terminology} Graph learning comprises several regimes. The most common ones are \emph{graph-level} and \emph{node-level predictions} (i.e., classification or regression). In the former, we are given a training set of graphs and aim to train a GNN to make meaningful graph-level predictions outside this training set. In the latter, we instead seek to make predictions for nodes in a given graph or set of graphs; the setup here is either \emph{transductive} or \new{inductive}. In the transductive setting, we are given a single graph with a subset of the nodes being the training set, and we aim to train a model to make correct predictions for the nodes outside this training set. In the inductive setting, we are given a training set of graphs with node (class-)labels and aim to train a model to make correct predictions for the nodes of unseen graphs. Similarly, we can define \new{edge-level} or \new{link prediction}. In addition, \emph{graph generation} aims to generate graphs modeled to a given data distribution proxied via a training dataset.

\paragraph{Related work} One of the first efforts towards more principled benchmarking of GNNs was taken by \citet{Dwi+2020}, who proposed a suite of real and synthetic graphs spanning a variety of node-, edge-, and graph-level tasks as well as an attempt to standardize evaluation protocols. However, the majority of the tasks either have a graph structure \emph{superimposed} on the original dataset (such as graphs extracted from vision datasets like \textsc{CIFAR10} which are long solved in the vision community) or focus on small synthetic graphs with a saturated performance. Another limiting factor is the strongly suggested model size below 500k parameters that was supposed to test models' \emph{inductive biases}. While reasonable for the state of graph learning in 2020, such a manually set parameter count ceiling makes little sense in modern deep learning where scaling laws suggest model capabilities grow with both dataset size and parameter count~\citep{hoffmann2022training, schaeffer2023are, wei2022emergent}.

Soon after, \citet{hu2020ogb} released the \new{Open Graph Benchmark} (\textsc{OGB}), a comprehensive suite of datasets encompassing various domains, tasks, and graph distributions. The authors proposed to gather results in a centralized, publicly visible leaderboard. The submission system requires researchers to provide test results, the corresponding validation performance, the number of learnable parameters, and some information about the tuning procedure. This effort goes in the direction of more informative and standardized benchmarking practices. Nevertheless, many datasets in the suite address (such as 2D molecular graphs or academic citation networks) are still a far cry from transformative real-world applications. As we discuss later in \Cref{subsec:apps,subsec:graph_construction}, these graphs either fail to encode relevant information (e.g., 3D spatial arrangements of atoms) or induce a structural inductive bias that is of unclear advantage for downstream generalization performance. While we note that some (large-scale) more impactful benchmarks are exposed by \textsc{OGB}, the research community has focused on them with relatively lower priority. This is likely due to the inherent difficulty of scaling more sophisticated and expressive architectures to larger graphs or the interest drawn by more specific settings, such as heterophilic networks, not generally covered by \textsc{OGB}.

\citet{Dwi+2022} proposed benchmark datasets to assess the long-range capabilities of GNNs, also transforming computer vision datasets into graph datasets, empirically showcasing the benefits of \new{graph transformers} (GTs)~\citep{Mue+2023} over MPNNs. However,~\citet{Toe+2024} have shown that the reported performance gap of graph transformers on these tasks is overestimated due to suboptimal hyperparameter choices, showcasing improper benchmarking practices. In addition, ~\citet{Err+2020} proposed a more meaningful evaluation protocol for GNNs; however, their efforts primarily focused on the small datasets from~\citet{Mor+2020}.

In 2D graph generation, many papers still evaluate on \textsc{QM9}~\citep{wu2018moleculenet} or \textsc{Zinc250k}~\citep{gomez2018automatic} even though these datasets are regarded as solved, i.e., most state-of-the-art models obtain near-perfect performance. In addition, the widely used \textsc{SPECTRE} benchmark~\citep{martinkus2022spectre} is also saturated, and results are not consistently reported across papers.

See~\cref{ext} for an extended discussion of related work.

\section{Missing transformative real-world applications and supporting benchmarks}
\label{subsec:apps}

We believe that the graph learning community has not yet identified benchmarks showcasing transformative real-world applications that genuinely exploit the benefits of machine learning on graphs. Unlike the computer vision or natural language domains, graph learning has no \say{natural} application areas, as graphs usually abstract other data modalities featuring more or less evident relational structures. 

In the past, graph learning primarily focused on benchmarking newly developed GNN architectures on datasets stemming from specific applications.  The molecular domain, e.g., predicting properties of small 2D molecular graphs~\citep{hu2020ogb,Mor+2020} has been an area of particular interest. Meanwhile, meaningful small 2D molecular graphs only cover minor, niche sub-fields in chemistry or drug discovery, where it is more natural to relate a 3D structure and a property evaluated at a quantum mechanical level of theory. In addition, transforming raw chemical data obtained from, e.g., experiments or quantum mechanical calculations into 2D molecular graphs can be time-consuming; it often results in the loss of important information and, thus, fails to capture the relationship between spatial atomic arrangements and properties.

In addition to challenges in supervised graph learning, similar issues arise in graph generation. Most papers benchmark their methods using 2D molecular graph generation~\citep{vignac2023digress}. However, 3D point clouds might be better suited and preferred by domain experts for such tasks, as the geometric structure of molecules is crucial for real-world applications, such as molecular docking or fragment linking \cite{igashov2024equivariant,schneuing2024structure}. For example, despite its prominence, DiGress~\citep{vignac2023digress}---one of the most cited works in graph generation over the past two to three years---has seen few practical follow-ups. Notably, most citations serve as background rather than extensions of their ideas. The utility of generating structured data remains unclear---with evaluating the quality of generated graphs without ground-truth data being one of the key challenges~\citep{Handa2023}---leaving this field without a clear application-driven focus. As a result, critical topics, such as generating graphs with strong structural constraints or scaling methods to large graphs, receive limited attention.

\paragraph{Suggested remedies} The community should shift focus from smaller, less relevant 2D molecular benchmarks to problems naturally represented as graphs. One promising area is {combinatorial optimization}, where graphs encode problem instances, such as in the vehicle routing problem~\citep{Tot+2002} or bipartite graphs in integer-linear programming~\citep{Schrijver86}, as discussed in~\citet{Cap+2021}. Combinatorial optimization benchmarks offer distinct advantages: (1) clear real-world applications, (2) easy generation of large datasets, and (3) ideal testbeds for studying size generalization.

Beyond combinatorial optimization, other high-potential areas for GNNs include {satisfiability solving}~\citep{Bie+2021}, {recommender systems}~\citep{Wu+2022}, {social networks}~\citep{New2003}, and {power-flow networks}~\citep{Owe+2020}. Projects like RelBench~\citep{robinson2024relbench} and 4DBInfer~\citep{dbinfer} demonstrate GNNs' utility in automating machine learning on {relational databases}, while TpuGraphs~\citep{tpugraphs} highlights their potential in {computer systems}. GNNs are also effective in {automated chip design}, such as in AlphaChip, where reinforcement-learning-based models leverage netlist embeddings (a hypergraph of circuit components and their connections)~\citep{mirhoseini2021graph, mirhoseini2021graphaddendum}.

Industrial datasets like social networks often involve sensitive data, limiting accessibility. Better {anonymization methods}, such as generating anonymous graphs similar to real-world data~\citep{yoongraphgen23}, could address this. While GNNs have been used for de-anonymization~\citep{cretu2022interaction}, anonymized graph generation remains an open challenge. 

Graph generation also holds promise for design-related tasks, mainly through diffusion models~\citep{liu2024graph}. For example, these models create heat maps for sampling solutions to combinatorial problems~\citep{li2024distribution,sun2023difusco}. While competitive, further work is needed to improve efficiency and understand GNN capabilities in this context. We hypothesize that for any prediction task $p(y|x)$, the conditional generation counterpart $p(x|y)$ is also valuable, provided $y$ is easy to evaluate or $p(y|x)$ is robust enough to assess generated samples reliably. By prioritizing combinatorial optimization and graph-representable problems, the community can advance theoretical insights and practical applications, providing a more straightforward path to real-world impact.

\section{Graphs are not necessarily constructed in a meaningful way}
\label{subsec:graph_construction}

As discussed above, graphs are higher-level abstractions of real-world phenomena or observables featuring relational structure. Hence, their effectiveness in tackling a specific task will inherently depend on how they are constructed and whether the relational information they encode predicts the problem~\citep{halcrow2020grale}. However, commonly adopted graph learning benchmarks often do not consider the meaningfulness, relevance, and completeness of the proposed constructed graphs; in fact, they sometimes either represent unsuitable formalisms for the data modality at hand, fail to encode important information, or do not correlate with the considered learning targets. We provide some examples in the following.

A first exemplary case is that of the \textsc{PascalVOC-SP} and \textsc{COCO-SP} datasets~\citep{Dwi+2022}. Their graphs encode coarse-resolution images, with rag-boundary edges drawn to connect super-pixels corresponding to segmented regions. However, this modeling choice is not grounded on any theoretical or empirical justification. As such, it is unclear whether modeling images as graphs in this way is helpful for object detection or the vision domain in general.

Spatiotemporal datasets, such as traffic networks, e.g., \textsc{PEMS-BAY} and \textsc{METR-LA} ~\citep{li2017diffusion} or air quality measurements. e.g., \textsc{AQI} \citep{zheng2015forecasting}, rely on sensor readings taken at various locations. Subsequently, a thresholded Gaussian kernel is applied to the pairwise distances between these sensor locations to construct the graph structure, 
introducing structure to an otherwise fully connected weighted graph by imposing a threshold to decide which connections are retained. While this preprocessing step provides a relational structure that facilitates using graph-based methods, it is fundamentally arbitrary and may misrepresent the system's dynamics. For instance, the choice of the threshold value is often heuristic, potentially omitting meaningful connections, e.g., emphasizing short-range interactions, which may or may not be the right choice for the problem. This highlights the need for more principled, data-driven methodologies for constructing spatiotemporal graphs.

Again, another relevant example is represented by the widely adopted \textsc{Zinc} benchmark~\citep{Dwi+2020}. \textsc{Zinc} contains small molecular graphs, whereby nodes represent atoms and edges the chemical bonds between them. This form of relational structure captures natural chemical information. Still, nodes and edges are attributed solely to the type of atoms and chemical bonds they represent, missing encoding important structural information such as the 3D atom coordinates and the SMILES-derived features that are easily obtained via software packages such as RDKit~\citep{Lan+2016}.

In some other compelling cases, relational information can be natural to consider but not necessarily informative to solve the prediction task at hand. In particular, this issue has been studied in recent work by~\citet{bechler-speicher2024graph}. The authors show settings where MPNNs overfit to spurious correlations in the structure, whereas set-based models~\citep{Zha+2017b} only process node features and exhibit better generalization performance. Exemplary settings of this kind are those of citation networks, where nodes represent scientific articles connected by edges whenever one cites the other. This form of relational structure is exceptionally reasonable but not necessarily predictive for any task instantiated on these graphs. That is, textual similarity in the content could, e.g., better correlate with article category than simply patterns of citations.

\paragraph{Suggested remedies} Virtually any real-world phenomenon and system can potentially be modeled as a \say{graph}~\citep{velickovic2023everything}, but this does not imply that any choice of relational structure is equally relevant or predictive or that a relational framework is a convenient modeling choice. Benchmarks should be designed in a way that accounts for these aspects systematically and quantitatively, openly and structurally considering the motto: 

\begin{center}
\emph{\say{Not everything that could be modeled as a graph should be modeled as a graph.}} 
\end{center}

When proposing a new benchmark, authors should discuss the advantages of adopting a \say{relational} modeling framework, articulating the advantages expected from processing data framed in graphs w.r.t.\ other possible modalities. In addition, they should discuss the choice of node and edge features and the rationale for how edges are determined in the first place. Crucially, authors should not only illustrate \emph{how} graphs are constructed but expand on \emph{why} the chosen approach is expected to be advantageous for the prediction task at hand. 

Quantitatively, we advise that benchmarks should always be accompanied by (adequately tuned) baselines such that comparisons with them will allow us to underscore the advantages of the considered structural information on generalization performance. Concretely, benchmarks should always report the performance of baselines that only process unstructured sets of node features,\footnote{These could be, for example, instantiated as DeepSets~\citep{Zha+2017b} or transformer-based architectures~\citep{Mue+2023}.} or graphs whose connectivity is obtained solely from the similarity thereof. Benchmark guidelines should explicitly promote the quantifying performance of proposed approaches in relative terms w.r.t.\ these.

\section{Bad benchmarking culture}
\label{sec:culture}

We believe inadequate benchmarking culture significantly hinders the graph learning community, irrespective of impactful applications (see~\Cref{subsec:apps}) or the usefulness of underlying graphs (see~\Cref{subsec:graph_construction}). While poor benchmarking exists across machine learning~\citep{Her+2024}, it is particularly problematic in graph learning. Even for standard datasets~\citep{Mor+2020}, inconsistent evaluation protocols and dataset splits result in highly variable performance reports (see~\Cref{subsec:enyzmes}), with some papers overestimating performance by reporting validation metrics~\citep{Err+2020}. Small datasets like \textsc{Mutag}~\citep{Mor+2020}, with only 188 graphs, lead to large standard deviations and unreliable comparisons, while some suffer from misclassifications or insufficient class representation~\citep{li2023graphcleaner, Platonov2023ACL}.

Newly proposed architectures are often unfairly compared to outdated baselines, with hyperparameters fine-tuned on a small number of datasets but not for baselines. Theoretically motivated GNNs~\citep{Mar+2019, Mor+2019} frequently claim inflated performance gains by avoiding comparisons with state-of-the-art models.

The community often overlooks the relevance of minor improvements. For instance, \textsc{Zinc}~\citep{Dwi+2020} tasks can be easily solved with standard chemoinformatics tools~\citep{Lan+2016}, yet incremental improvements on such benchmarks are often highlighted. Additionally, limited molecular and material modeling domain knowledge prevents meaningful task understanding. For example, current state-of-the-art models often ignore critical relationships between 3D structure and molecular properties.

In 2D graph generation, datasets like \textsc{QM9}~\citep{wu2018moleculenet} and \textsc{Zinc250k}~\citep{gomez2018automatic} dominate despite near-perfect performance. More robust benchmarks, such as \textsc{MOSES}~\citep{polykovskiy2020molecular} and \textsc{GuacaMol}~\citep{brown2019guacamol}, remain underutilized due to high computational demands. Benchmarking inconsistencies, such as differing dataset splits~\citep{siraudin2024cometh}, inappropriate reliance on novelty for \textsc{QM9}~\citep{vignac2021top}, and inconsistent FCD reporting further exacerbate the issue. Current benchmarks emphasize unconditional generation, whereas real-world applications require conditional generation, which remains underexplored due to oversimplified tasks and strong baselines~\citep{tripp2021fresh}.

Beyond molecules, benchmarking for graph generative models is even less standardized. Some studies rely on limited datasets like \textsc{Cora} or the \textsc{SPECTRE} benchmark~\citep{martinkus2022spectre}, focusing on specific graph types but often omitting metrics like VUN and error bounds. Benchmarking for large graphs faces additional challenges due to the scarcity of practical datasets and even poorer standardization practices.

\paragraph{Suggested remedies}
To address these challenges, the graph learning community must develop practical tasks and robust evaluation frameworks. Unlike LMsys Arena's ELO-based evaluation~\citep{zheng2023judging}, graph learning lacks trusted benchmarks resistant to manipulation. While domain expertise poses challenges, creating expert-validated benchmarks can significantly improve model evaluation and adoption.

A Kaggle-like competition with hidden test sets at the NeurIPS benchmark track could realistically assess models across domains like molecular prediction and combinatorial optimization. Addressing data quality issues requires larger, domain-relevant datasets such as \textsc{ADMET Benchmark Group}~\citep{Swa+2023} or \textsc{PubChemQC PM6}~\citep{Nak+2020}, which provide diverse, real-world data. Multidisciplinary collaboration is essential for curating datasets and translating real-world problems into graph learning tasks~\citep{you2020design}.

For 2D molecule generation, benchmarks like \textsc{MOSES} and \textsc{GuacaMol} should replace outdated ones like \textsc{QM9} and \textsc{Zinc250k} for serious evaluations. Future efforts must focus on computational efficiency and improved benchmarks such as \textsc{SPECTRE}, incorporating larger datasets with diverse structural properties. Evaluations must include error bars and report ratios and prioritize combined metrics like MMD and VUN. We advocate for new benchmarks extending existing frameworks to effectively evaluate diverse, complex structures.

\section{Implication: No true foundation model exists for graph learning}\label{sec:gfm}

In deep learning, large pre-trained foundation models~\citep{llama3,gemini} that unify multiple modalities (e.g., text, images, video, audio) excel at predictive and generative tasks, reshaping research and industry. However, similarly impactful \new{graph foundation models} (GFMs) are yet to emerge. Domain-specific graph-based models have appeared recently~\citep{mao2024position} for tasks such as node classification~\citep{zhao2024graphany}, neural algorithmic reasoning~\citep{ibarz2022a}, knowledge graph reasoning~\citep{galkin2023ultra,galkin2024ultraquery}, and molecular property prediction~\citep{kla+2024,sypetkowski2024on}. Yet, their performance often shows only marginal gains over standard supervised GNNs~\citep{zhao2024graphany,kla+2024,chen2024textspace}.

As argued in~\cref{subsec:apps,subsec:graph_construction,sec:culture}, training on small datasets or academic tasks without rigorous evaluations hampers progress in graph learning. Additional challenges include: (1) differing symmetries and expressivity requirements across tasks (e.g., labeling trick GNNs~\citep{labeling_trick,nbfnet} excel in link prediction but not node- or graph-level predictions); (2) learning representations for graphs with varying scales and feature spaces necessitates new strategies for \emph{graph tokenization} and defining a universal \emph{graph vocabulary}~\citep{mao2024position}; (3) graph data availability is orders of magnitude smaller than text data, and a \emph{token} for graphs lacks clear definition; (4) limited commercially relevant GFM applications, as discussed in~\cref{subsec:apps}.

\paragraph{Suggested remedies} Despite these challenges, GFMs and robust real-world graph benchmarking are critical for advancing graph learning alongside progress in other deep learning areas. We propose shifting from \emph{one model for one dataset} to \emph{one model for all datasets} to provide a comprehensive view of model performance across diverse graphs. For example, instead of training a separate model for each task in a five-task benchmark, training one model for all tasks is preferable. For particularly non-trivial setups (e.g., combining classification with regression), we suggest an \emph{encoder-processor-decoder} approach~\citep{battaglia2018relational,ibarz2022a}: pre-train a unified backbone model and fine-tune task-specific encoders and decoders. Finally, we advocate for creating large-scale, high-quality datasets of diverse graph structures (e.g., sparse, dense, homophilic, heterophilic, directed, multi-relational), addressing data gaps with synthetic data~\citep{palowitch2022graphworld}, and ensuring data decontamination by excluding known test sets from pre-training corpora.

\section{Alternative views}

Fields adjacent to graph learning, such as \emph{geometric deep learning} (GDL)~\citep{Bron+2017,bronstein2021gdlbook}, are thriving and achieving remarkable successes. GDL has driven advancements in structural biology~\citep{abramson2024AF3,Jum+2021,Tow+2021} and materials science~\citep{merchant2023scaling,Rei+2022,Zeni2025}. It also underpins state-of-the-art interatomic potentials for atomistic simulations at first-principles accuracy~\citep{batatia2022mace,batzner2023,Gasteiger2022c,hu2021forcenet,musaelian2023learning,osti_1784359,schutt2017schnet,schutt2021equivariant,Simeon2023,thomas2018tensor,Zaverkin2024b}, including the trend toward universal interatomic potentials~\citep{Batatia2023,Chen2022AUG,Devereux2020,Smith2017,Kovacs2023}. GDL focuses on 3D objects, addressing symmetries and invariances inherent to input data. Its success stems from tackling real-world applications, such as the 2024 Nobel Prize-winning advancements in protein design, meaningful benchmarks based on real-world data, and foundational models for practical tasks. While graph structure can be useful in GDL, it is often secondary, with inputs treated as fully connected graphs (within a cutoff distance) or point clouds. In contrast, classical graph learning on molecules involves 2D objects where graph structure and permutation symmetry are the primary \emph{inductive biases} (often noisy, see \cref{subsec:graph_construction}) and lacks the rich 3D information critical in GDL.

\section{Empirical evidence}

Here, we support our claims made in the previous four sections with empirical evidence\footnote{Our code is available at \url{https://anonymous.4open.science/r/PP-Benchmarks-82C1}.}.

\subsection{Graphs not necessarily constructed in a meaningful way}

\begin{table}[ht!]
\centering
\setlength{\tabcolsep}{0.4em}
\caption{Comparison of different GNNs over \textsc{OGB} datasets, when using DeepSets (no graph), the original graph (Orig.) and a fixed expander graphs (Cayley).}
\label{tab:empty_cayley_orig_eval}
\vskip 0.15in
\resizebox{0.9\linewidth}{!}{%
\begin{tabular}{lcccc}
\toprule
\textbf{Model} & \textsc{molhiv} & \textsc{molbbbp} & \textsc{molbace} \\
\midrule
Deepset (Empty) & \text{63.78}$_{\pm \text{1.05}}$ & \text{64.90}$_{\pm \text{0.72}}$ & \text{51.76}$_{\pm \text{2.85}}$ \\
\midrule
GraphConv  Orig. & \textbf{\text{68.24}$_{\pm \text{1.77}}$} & \textbf{\text{64.11}$_{\pm \text{4.50}}$} & \textbf{\text{63.18}$_{\pm \text{4.56}}$} \\

GraphConv Cayley & \textbf{\text{67.91}$_{\pm \text{0.75}}$} & \text{61.60}$_{\pm \text{4.48}}$ & \text{56.94}$_{\pm \text{7.50}}$ \\
\midrule
GIN  Orig. & \textbf{\text{69.65}$_{\pm \text{2.58}}$} & \textbf{\text{66.73}$_{\pm \text{1.27}}$} & \text{53.44}$_{\pm \text{4.52}}$ \\

GIN Cayley & \textbf{\text{68.61}$_{\pm \text{1.40}}$} & \text{58.35}$_{\pm \text{4.01}}$ & \textbf{\text{56.94}$_{\pm \text{12.40}}$} \\
\midrule
GAT  Orig. & \textbf{\text{67.21}$_{\pm \text{1.30}}$} & \textbf{\text{66.62}$_{\pm \text{1.14}}$} & \text{53.21}$_{\pm \text{1.34}}$ \\

GAT Cayley & \textbf{\text{67.80}$_{\pm \text{3.45}}$} & \text{60.31}$_{\pm \text{2.47}}$ & \textbf{\text{62.75}$_{\pm \text{4.76}}$} \\
\bottomrule
\end{tabular}
}
\end{table}

In~\cref{subsec:graph_construction}, we raised concerns about the lack of correlation between graph structures in commonly used benchmarks and the intended learning targets. In this subsection, we provide empirical evidence to support this claim further. Recently, \citet{deac2022expandergraphpropagation, wilson2024cayleygraphpropagation} proposed a message-passing scheme in which, during every odd layer, the original graph is disregarded in favor of propagating information through a fixed-structure expander graph—specifically, a Cayley graph. Ablation studies presented in \citet{wilson2024cayleygraphpropagation} on multiple \textsc{TUDataset} benchmarks showed that using the Cayley graph exclusively, without incorporating the original graph at any layer, sometimes improved performance. This finding is striking, as the Cayley graph does not inherently encode task-relevant information.
These results align with the observations of~\citet{bechler-speicher2024graph}, who showed that making graphs more regular consistently improved performance. To further substantiate these findings, we replicate these experiments using the \textsc{OGB} graph-level benchmarks, strengthening the evidence for these observations. We also evaluate a DeepSets~\cite{zaheer2018deepsets} baseline, where we drop the graph structure from the data, and therefore, the MPNN acts on an empty graph. Due to space limitations, the Appendix provides all the experimental details, including dataset information and tuned hyper-parameters.

The ROC-AUC scores average over $3$ random seeds, are summarized in Table~\ref{tab:empty_cayley_orig_eval}. The best-performing model within the standard deviation range is marked in bold for each dataset. Across 5 out of 9 experiments, the non-informative regular Cayley graph outperformed or matched the performance of the original graph. Notably, for the \textsc{molbbbp} datasets, training with GraphConv achieved the highest AUC-ROC when the graph structure was completely dropped.

\subsection{Reassessing simple baselines on \textsc{PCQM4Mv2}}
\begin{table}[!htb]
\caption{Evaluation results on the validation split of PCQM4Mv2.}
\label{tab:pcqm}
\centering
\setlength{\tabcolsep}{0.4em}
\resizebox{0.9\linewidth}{!}{%
    \begin{tabular}{clcc}
    \toprule
    & \textbf{Model} & \textbf{Val. MAE} & \textbf{\#Param.} \\
    \midrule
    \multirow{2}{*}{\rotatebox{90}{GNNs}}
    & GCN \cite{hu2ogb} & 0.1379 & 2.0M \\
    & GINE \cite{hu2ogb} & 0.1195 & 3.8M \\
    %& GCN+VN \cite{hu2ogb} & 0.1153 & 4.9M \\
    %& GIN+VN \cite{hu2ogb} & 0.1083 & 6.7M \\
    \midrule
    \multirow{5}{*}{\rotatebox{90}{Transformers}}
    & TokenGT \cite{kim2022pure} & 0.0910 & 48.5M \\
    & GRPE \cite{park2022grpe} & 0.0890 & 46.2M \\
    & Graphormer \cite{shi2022benchmarking} & 0.0864 & 48.3M \\
    & GPS \cite{rampavsek2022recipe} & 0.0858 & 19.4M \\
    & ET \cite{mueller2024towards} & 0.0832 & 16.8M \\
    \midrule
    \multirow{2}{*}{\rotatebox{90}{Ours}} 
    & GINE & 0.0913$_{\pm\text{0.0002}}$ & 22.7M \\
    & GINE+RWSE & 0.0898$_{\pm\text{0.0001}}$ & 22.7M \\
    \bottomrule
    \end{tabular}
}

\end{table}

In \Cref{sec:culture}, we discussed problematic practices of empirical evaluations of novel GNN architectures. One common issue is the citation of old, outdated reference results to quantify the performance improvements of new architectures over simpler baselines. Often, these baseline results suffer from suboptimal hyper-parameters and are cited as-is for many years without reevaluation. As a consequence, the performance gains of newer architectures are commonly overestimated. 

Here, we demonstrate this issue on the commonly used \textsc{PCQM4Mv2} dataset \citep{hu2ogb}, among the few large-scale datasets for graph-level learning tasks and is particularly popular for demonstrating performance improvements of graph transformers over simpler MPNNs. Experimental evaluations on this dataset typically cite the results for GCN \citep{Kip+2017} and GINE \citep{xu2018powerful, hu2020pretraining} that were initially reported on the leaderboard of the 2021 \textsc{OGB-LSC} competition \citep{hu2ogb} to represent standard MPNNs. 
These results suggest a validation MAE of around 0.12, while graph transformers commonly achieve MAEs below 0.09, indicating a substantial error reduction of over 25\%. 

We aim to reevaluate this performance difference by reassessing the reference results for standard message-passing GNNs. Specifically, we measure GINE's performance after re-tuning hyper-parameters for a larger 20-layer model with approximately 20 million parameters. We base our experiments on the same GINE architecture with edge features used to obtain the original results.
We make minor adjustments to align the setup with current deep learning practices used for transformers. Full details, hyper-parameters, and tuning budgets are reported in~\cref{app_exp}. We report performance for GINE models with and without additional RWSE node features \citep{dwivedigraph}, also used by graph transformers such as GPS \citep{rampavsek2022recipe}.

\Cref{tab:pcqm} provides the results of our evaluation. Even without additional RWSE features, the error of GINE drops by over 20\% to 0.0913 simply by tuning the model configuration. When using additional structural features, the performance improves further to 0.0898, which is competitive with several graph transformer baselines. We do not claim that additional tuning could not further improve the graph transformers' results. Instead, our results show how brittle empirical evaluations of GNNs generally are and that the numbers reported throughout the literature often do not capture the actual progress of model capability or the lack thereof.

\subsection{The meaningfulness of architectural changes}
\begin{table}[htp!]
    \centering
    \setlength{\tabcolsep}{0.4em}
    \caption{Comparison of baseline GNNs with and without architectural modifications on heterophilous datasets.}
    \label{tab:heterophilic}
    \resizebox{.9\linewidth}{!}{%
    \begin{tabular}{clccc}
        \toprule
        & \textbf{Model} & \textsc{roman-empire} & \textsc{amazon-ratings} & \textsc{minesweeper} \\ %& \textbf{tolokers} & \textbf{questions}\\
        \midrule
        \multirow{2}{*}{\rotatebox{90}{GCN}} &
        Re-evaluated & 44.41$_{\pm 0.81}$ & 44.30$_{\pm 0.52}$ & 72.90$_{\pm 1.29}$\\ %& 69.97$_{\pm 1.29}$ &\\
        & Reported & 73.69$_{\pm 0.74}$ & 48.70$_{\pm 0.63}$ & 89.75$_{\pm 0.52}$\\ %& 83.64$_{\pm 0.67}$ & 76.09$_{\pm 1.27}$ \\
        \midrule
        \multirow{2}{*}{\rotatebox{90}{SAGE}} &
        Re-evaluated & 80.80$_{\pm 0.52}$ & 43.35$_{\pm 0.80}$ & 83.76$_{\pm 0.71}$\\
        & Reported & 85.74$_{\pm 0.67}$ & 53.63$_{\pm 0.39}$ & 93.51$_{\pm 0.57}$\\ %& 82.43$_{\pm 0.44}$ & 76.44$_{\pm 0.62}$ \\
        \midrule
        \multirow{2}{*}{\rotatebox{90}{GAT}} &
        Re-evaluated & 51.05$_{\pm 0.90}$ & 44.52$_{\pm 0.48}$ & 74.37$_{\pm 0.94}$\\
        & Reported & 80.87$_{\pm 0.30}$ & 49.09$_{\pm 0.63}$ & 92.01$_{\pm 0.68}$\\ %& 83.70$_{\pm 0.47}$ & 77.43$_{\pm 1.20}$ \\
        \midrule
        % & Rel. Avg. Gain & +21.35 & +6.42 & +14.75\\
        & Avg.\ \% Gain & +43.56\% & +14.63\% & +19.49\%\\
        \bottomrule
    \end{tabular}
    }
\end{table}

In addition, in response to~\Cref{sec:culture}, we further exemplify the GNN evaluations' brittleness in the node-prediction setting. \citet{Platonov2023ACL} proposed a set of heterophilous graph datasets to evaluate the performance of various GNNs, including both baseline and heterophily-specific GNNs. In their experiments, the authors used the official implementations of the heterophily-specific GNNs. However, they modified the baseline GNNs by adding a two-layer MLP after each neighborhood-aggregation layer. While mentioning that this architectural enhancement significantly improved baseline performance, the authors did not explore its impact further.  

This enhancement raises several critical concerns regarding the validity of the comparisons: (1) No parameter budget was enforced, potentially leading to models of varying capacities. (2) The evaluation of the baseline GNNs followed a uniform protocol, whereas heterophily-specific GNNs were assessed using their respective codebases, which may incorporate diverse architectural choices and introduce unfairness. However, most important of all, (3) the significance of the specific baseline GNN protocol---including the two-layer MLP after each graph neighborhood aggregation, a linear encoder, a linear decoder, and GeLU activation---was acknowledged but not thoroughly analyzed, leaving its contribution to performance improvement unclear.  

These issues render the performance comparisons in \citet{Platonov2023ACL} less meaningful. Although the proposed datasets may serve as valuable benchmarks for heterophilous graphs, their utility cannot be conclusively determined without an evaluation protocol. This underscores the necessity of such a protocol (see \cref{sec:culture})---even the most promising benchmarks require well-defined evaluation guidelines to assess their quality reliably. 

To validate these concerns, we re-evaluated the baseline GNNs (GCN \cite{Kip+2017}, SAGE \cite{Ham+2017}, and GAT \cite{velivckovic2017graph}) using a fresh codebase that adhered to the hyper-parameters reported in \cite{Platonov2023ACL} but excluded specific architectural modifications. Specifically, our evaluation omitted the linear encoder, the two-layer MLP after each aggregation layer, and the linear decoder and replaced GeLU activations with the standard ReLU.  

As shown in \cref{tab:heterophilic}, these architectural changes introduced in \citet{Platonov2023ACL} resulted in significant average baseline performance gains of +43.56\%, +14.63\%, +19.49\% on the roman-empire, amazon-ratings, minesweeper datasets, correspondingly. This analysis does not question the validity of the proposed benchmarks but highlights the critical need for accompanying evaluation protocols. Such protocols should include a fixed model size limit to ensure fair parameter budgets and clear guidelines on allowable architectural modifications across all GNN layers. 

\subsection{Multi-task pre-training with encoder-processor-decoder}

\begin{table}[h!]
\caption{Test performance on the upstream datasets, both trained in a single-task (ST) and multi-task (MT) setting, as well as random baselines (RD), on a single random seed.}
\label{tab:fm_upstream}
\centering
\resizebox{0.9\linewidth}{!}{%
    \begin{tabular}{clcccc}
    \toprule
& \multirow{2}{*}{\textbf{Model}} & \textsc{COCO-SP} & \textsc{MalNetTiny} & \textsc{PCQM4Mv2} \\ \cmidrule{3-5}
 & & F1 $\uparrow$ & Acc. $\uparrow$ & MAE $\downarrow$ \\\midrule
     \multirow{2}{*}{\rotatebox{90}{RD}}
    & MPNN & 0.0002 & 23.10 & 5.2340 \\
    & GT & 0.0005 & 19.60 & 5.2483 \\
    \midrule
    \multirow{2}{*}{\rotatebox{90}{ST}}
    & MPNN &	0.0817 & 81.80 & 0.1104	 \\
    & GT & 0.2947 & 81.90 & 0.1009 \\
    \midrule
    \multirow{3}{*}{\rotatebox{90}{MT}}
    & Empty & 0.0119 & 20.00 & 0.3915 \\
    & MPNN &	0.0413 & 83.20 & 0.1363 \\
    & GT & 0.1137 & 88.90 & 0.1441 \\
    \bottomrule
    \end{tabular}
}
\end{table}

In this section, as suggested in~\Cref{sec:gfm}, we run a series of experiments to investigate multi-task pre-training/fine-tuning using an encoder-processor-decoder framework. For small molecules, similar settings have been explored recently in \citet{kla+2024, sypetkowski2024on,frasca2024towards}. Here, we also want to study a cross-domain setting with data from vision, function-call graphs, large molecules, and social networks.
The aim is to gather an initial signal on the suitability of this architectural pattern when pre-trained on a mix of vastly different graph tasks, even on a relatively small scale. As highlighted in~\Cref{sec:gfm}, a large-scale, curated pre-training corpus is currently lacking, and we believe that positive results from our experiments could catalyze the community's efforts in building such a corpus, accompanied by standardized pre-training setups and evaluation procedures.

\paragraph{Architectures, training, and evaluation}
We train domain-specific encoders (e.g., embedding atom and bond types in molecules) and task-specific decoder MLPs. For the processor network, we both evaluate an MPNN based on the GINE architecture \citep{Xu+2018} and a GT based on Graphormer \citep{ying2021transformers} with a soft attention bias and RWSE structural encodings \citep{dwivedigraph}.  For our experiments, we assemble two sets of datasets: the \textit{upstream mix} used for pre-training and the \textit{downstream mix} used for fine-tuning. We freeze the processor weights during fine-tuning and learn a new encoder and decoder. If the downstream task permits, we reuse the encoder from one of the datasets in the upstream mix. In addition, for each upstream or downstream dataset, we train baseline models with the same model architectures as our pre-trained models but trained from scratch in a single-task fashion. All experimental details are enclosed in~\Cref{app:gfm}.

\paragraph{Upstream mix}
Our upstream mix contains \textsc{PCQM4Mv2} \citep{hu2ogb}, \textsc{COCO-SP} \citep{Dwi+2022}, and \textsc{MalNetTiny} \citep{freitas2021malnet}.  These datasets are diverse in various aspects, such as the underlying application domain, the type of prediction task, and graph size and sparsity. As a result, we do not expect a strong transfer between any of these tasks during pre-training. Rather, we select this upstream mix to investigate whether our MPNN and GT models can learn \textit{general graph representations} useful for multiple, potentially unrelated tasks. To further support this assessment and quantify the benefits of learning a general-purpose processor network, we train an additional \say{empty graph baseline} in the multi-task setup, where the processor network is set to identity, and, hence, the graph structure is ignored. See \Cref{tab:fm_upstream} for our results. We observe that both the MPNN and the GT show non-trivial performance compared to the empty graph baseline and their randomly initialized (untrained) counterparts on all three tasks. On \textsc{COCO-SP} and \textsc{PCQM4Mv2}, they both fall short of their single-task baseline. Surprisingly, on \textsc{MalNetTiny}, we find that MPNN and GT improve over their respective single-task performance when trained in a multi-task setting.

\paragraph{Downstream mix}
Next, we evaluate how well the pre-trained multi-task models transfer to new downstream tasks. To this end, we construct a \textit{downstream mix} consisting of three datasets with various degrees of similarity to the upstream mix. In particular, we select \textsc{PascalVOC-SP} \citep{Dwi+2022}, \textsc{Peptides-Struct} \citep{Dwi+2022}, and \textsc{Stargazers} \citep{rozemberczki2020karateclub, Mor+2020}.\footnote{We detail the relation between \say{upstream} and \say{downstream} datasets in~\Cref{app:gfm}.} Here, we measure performance for a varying number of fine-tuning steps to assess whether the pre-trained models are more sample-efficient than their single-task counterparts; see~\Cref{fig:fm_downstream} for the results. 
We observe that pre-training is generally beneficial in the regimes of the fewest optimization steps, although to a degree that depends on the target dataset and chosen backbone. We observe strong in-domain transfer to \textsc{PascalVOC-SP} and strong cross-domain transfer to \textsc{Stargazers} on both pre-trained models. The results on \textsc{Peptides-struct} are less pronounced. While we observe slight transfer for the pre-trained GT, the pre-trained MPNN shows negative transfer at 3K and 10K steps.
%More specifically, on \textsc{PascalVOC-SP}, which shares both the domain (images) and the task (semantic segmentation) with \textsc{COCO-SP}, we observe strong performance already after roughly two epochs (1K steps), indicating effective transfer. On \textsc{Peptides-struct}, which shares its domain (molecules) with \textsc{PCQM4Mv2} but contains structurally different graphs and has different prediction targets, the GT benefits only marginally from pre-training. At the same time, the MPNN even indicates negative transfer at 3K and 10K fine-tuning steps, underperforming the MPNN baseline trained from scratch. MPNN and GT benefit from pre-training when transferred to \textsc{Stargazers}, a dataset from a `new' domain (social networks).

Overall, the above results suggest that MPNNs and GTs can learn general-purpose graph representations even when trained on data from different domains. These representations can often transfer effectively to in- and cross-domain tasks.

\section{Conclusion}

This paper highlights the need to rethink benchmarks and practices in graph learning. While GNNs have succeeded in many applications, current benchmarks often overlook real-world problems, focus too narrowly on specific data modalities, and lack consistent evaluation protocols or large-scale datasets for foundation models. We propose designing benchmarks that reflect real-world complexity, standardizing evaluations, and creating scalable datasets. These changes will help the graph learning community align with machine learning advancements and maintain impact and relevance.

\section*{Acknowledgements} 
We thank Erik Müller for crafting the figures and Petar Veličković for feedback on the paper.
AS, CM, and LM are partially funded by a DFG Emmy Noether grant (468502433) and RWTH Junior Principal Investigator Fellowship under Germany’s Excellence Strategy. AS performed this work as part of the Helmholtz School for Data Science in Life, Earth and Energy (HDS-LEE) and received funding from the Helmholtz Association of German Research Centres. BF is funded by the Clarendon scholarship. FF conducted this work while partly supported at the Technion by an Aly Kaufman and an Andrew and Erna Finci Viterbi Post-Doctoral Fellowship.

\bibliography{bibliography}
\bibliographystyle{icml2025}

\appendix

\newpage
\onecolumn

\section{Extended related work}\label{ext}

\textsc{GraphWorld}~\citep{palowitch2022graphworld} offered a synthetic perspective on graph benchmarking by showing that existing datasets cover a relatively narrow distribution of possible graphs and tuning common MPNN architectures is not indicative of their performance in other, less common domains. To alleviate the distribution issue, \textsc{GraphWorld} suggested generating synthetic graphs using stochastic block models~\citep{sbm} with more diverse connectivity patterns and probing GNNs on the synthetic datasets. Unfortunately, the dataset did not receive significant attention and adoption in the graph learning community partly due to the stated synthetic nature of the tasks.

In addition, \citet{velivckovic2022clrs} proposed \textsc{CLRS}, an algorithmic reasoning benchmark modeling the simulation of 30 classical algorithms as graph tasks such as node- or edge-level prediction and evaluates in a difficult size generalization setting. Algorithmic reasoning, in particular in the size generalization setting, receives some interest in the broader machine learning community~\citep{zhou2024algoreaso, zhou2024lengthgennotrobustly, mcleish2024transformers} and is arguably highly relevant to the study and advancement of the reasoning capabilities of neural networks in general. At the same time, algorithmic reasoning methods are typically benchmarked on synthetic tasks aimed at studying reasoning and learning capabilities in a controlled setting. There is no suitable replacement for high-quality, real-world benchmarks with direct downstream applications.

Furthermore, the benchmarking landscape for MPNNs remains constrained by the lack of large-scale and realistic graph datasets, particularly in domains like social networks. Commonly used datasets such as \textsc{Reddit}~\citep{Ham+2017} and \textsc{Flickr}~\citep{zeng2019graphsaint} are often cited as representative of real-world social networks. However, these datasets fail to capture key characteristics of actual social networks, such as high-degree hubs and dense community structures, as their average node degree is significantly lower. This discrepancy makes them poor representatives for large, realistic graphs. Similarly, in social network-based datasets such as the Twitter retweet-induced subgraph dataset~\citep{ribeiro2017like}, it is unclear whether the features and adjacency relationships of the sampled subgraph align with those of the full graph. Moreover, the structure of these subgraphs is constructed using a random walk-based crawler on the original graph. This sampling process further reduces the average node degree, making the dataset less representative of large-scale graphs, similar to \textsc{Reddit} and \textsc{Flickr}. We note here that the unavailability of real-world social network graph data is likely due to factors outside our field's control, e.g., privacy concerns and commercial relevance.

Another important aspect of many real-world graphs is their inherently dynamic nature. That is, nodes and edges and their features change over time. This aspect is often neglected in many datasets, including social networks. Recent efforts have introduced valuable benchmarking suites for learning on temporal graphs, e.g., the \emph{Temporal Graph Benchmark} (\textsc{TGB})~\citep{huang2023temporal,gastinger2024tgb}. Interestingly, on these temporal datasets, researchers have found that overlooked baselines and simple heuristics can be particularly predictive and outperform more sophisticated temporal GNNs~\citep{gastinger2024tgb,gastinger2024history}. This puts in question the relevance of some of the proposed benchmarks and the significance of the progress made by the community.

%In 2D graph generation, many papers still evaluate on \textsc{QM9}~\citep{wu2018moleculenet} or \textsc{Zinc250k}~\citep{gomez2018automatic} even though these datasets are regarded as solved, i.e., most state-of-the-art models obtain near-perfect performance. In addition, the widely used \textsc{SPECTRE} benchmark~\citep{martinkus2022spectre} is also saturated, and results are not consistently reported across papers.

\section{Additional experimental details}\label{app_exp}

Here, we provide additional experimental details and results.

\subsection{Graphs not necessarily constructed in a meaningful way}
\begin{table*}[h!]
\centering
\setlength{\tabcolsep}{0.7em}
\caption{Best hyper-parameters in the format num.\ of layers, width, batch size, learning rate, dropout.}
\label{tab:param_table_empty_cayley_orig_eval}
\begin{tabular}{lccc}
\toprule
\textbf{Model} & \textsc{ogbg-molhiv} & \textsc{ogbg-molbbbp} & \textsc{ogbg-molbace} \\
\midrule
GraphConv Orig. & $3$, $64$, $32$, $5e{-4}$, $0$ & $5$, $32$, $16$, $1e{-4}$, $0$ & $5$, $32$, $32$, $5e{-4}$, $0$ \\
GraphConv Empty & $3$, $64$, $32$, $1e{-4}$, $0$ & $3$, $64$, $32$, $5e{-4}$, $0$ & $3$, $32$, $16$, $5e{-4}$, 0.3 \\
GraphConv Cayley & $3$, $64$, $32$, $5e{-4}$, 0.3 & $3$, $64$, $32$, $5e{-4}$, $0$ & $5$, $64$, $32$, $5e{-4}$, $0$ \\
\midrule
GIN Orig. & $5$, $32$, $32$, $1e{-4}$, $0$ & $3$, $64$, $16$, $1e{-4}$, $0$ & $3$, $32$, $32$, $1e{-4}$, 0.3 \\
GIN Empty & $3$, $64$, $32$, $5e{-4}$, 0.3 & $3$, $32$, $32$, $5e{-4}$, $0$ & $3$, $32$, $32$, $1e{-4}$, 0.3 \\
GIN Cayley & $5$, $32$, $32$, $1e{-4}$, $0$ & $5$, $64$, $64$, $5e{-4}$, $0$ & $5$, $64$, $16$, $1e{-4}$, $0$ \\
\midrule
GAT Orig. & $5$, $64$, $32$, $5e{-4}$, $0$ & $3$, $64$, $16$, $5e{-4}$, $0$ & $5$, $32$, $32$, $1e{-4}$, 0.3 \\
GAT Empty & $3$, $32$, $32$, $1e{-4}$, $0$ & $3$, $32$, $32$, $5e{-4}$, $0$ & $5$, $32$, $32$, $1e{-4}$, 0.3 \\
GAT Cayley & $3$, $64$, $32$, $1e{-4}$, $0$ & $3$, $64$, $32$, $5e{-4}$, $0$ & $5$, $64$, $16$, $1e{-4}$, $0$ \\
\bottomrule
\end{tabular}
\end{table*}

For each model among GraphConv, GIN, and GAT, 
we tuned the learning rate in $\{10^{-3}, 5\cdot 10^{-3}\}$, number of layers in $\{3, 5\}$, dropout in $\{0, \text{0.3}\}$, hidden dimensions in $\{32, 64\}$, batch size in $\{16, 32\}$, early stopping  with patience of $50$ steps on the validation loss, and sum-pooling. We used ReLU activation and CrossEntropy loss.

For each dataset, we trained with seed $0$ over all the hyper-parameter configurations and selected the best performing configuration on the validation set, according to the ROC-AUC scores. We then trained each model with its selected configuration with seeds $1$ and $2$. Finally, we report the mean and standard deviation of the ROC-AUC scores over the test set over these $3$ seeds.

We consider the CGP propagation scheme from~\citet{wilson2024cayleygraphpropagation}, where for each model,  we utilize the Cayley graph in each layer and do not consider the original graph at all. For the DeepSet evaluation, we used the same architecture of GraphConv and fed it with empty graphs.

\subsection{Reassessing simple baselines on \textsc{PCQM4Mv2}}
We make minor changes to the layer configuration by using SiLU activation \citep{hendrycks2016gelu} instead of ReLU for improved gradient flow. We also replace the BatchNorm with LayerNorm and apply it \emph{after} the skip connection, similar to a standard transformer encoder layer. In each GINE layer, we use a 2-layer MLP as an update function with a hidden dimension of 1024 (double the embedding dimension (512)). After the final GINE layer, we apply sum-pooling followed by a 3-layer MLP, outputting a graph-level prediction.

We set the number of GINE layers to 20 and the latent embedding dimension to 512.
In preliminary experiments, we found larger models to be overfitting, so we fixed this model configuration with approximately 20 million trainable parameters. This is comparable to the model sizes used in the graph transformer literature and about five times larger than the GINE model used to obtain the original results.
Note that despite a model depth of 20 layers, we observed no performance degradation due to over-smoothing, trivially mitigated by following basic deep learning practices such as skip-connections and deep MLPs as update functions. It is known, of course, that these practices also prevent smoothing phenomena in transformers \citep{pmlr-v139-dong21a}, and the same holds for MPNNs.

We train with the L1-loss for one million gradient descent steps using a batch size of 512 with the Adam optimizer. The learning rate warms up linearly for the first $10^4$ steps and follows a cosine decay schedule for the remainder of training towards a minimum rate of $10^{-6}$.
No gradient clipping is used. We tune the remaining hyper-parameters through a grid search.
Specifically, we tune the learning rate in $\{2\!\cdot\!10^{-4}, 1\!\cdot\!10^{-4}, 5\!\cdot\!10^{-5}\}$, the dropout rate in $\{0, 1\!\cdot\!10^{-1}, 2\!\cdot\!10^{-1}\}$ and the weight decay in $\{0, 1\!\cdot\!10^{-1}\}$.
Tuning is done \emph{with} RWSE features, and we reuse the same configuration for GINE without RWSE.
Since the original validation split of \textsc{PCQM4Mv2} is used to compare models in the literature, we create a separate holdout set by sampling 10K graphs uniformly at random from the training data and use this set for model selection and hyperparameter tuning.
Each training run uses a single Nvidia H100 GPU and lasts approximately 8 hours.
In total, hyperparameter tuning consumed less than 200 H100 hours of computing.
The final hyper-parameters are provided in \Cref{tab:pcqm_hp}. For the final results reported in \Cref{tab:pcqm}, we average the performance over three runs with different random seeds and also provide the corresponding standard deviation, which is relatively low.
\begin{table}[h]
    \centering
    \caption{Hyper-parameters used for our evaluation of GINE on PCQM4Mv2.}
\label{tab:pcqm_hp}
    \begin{tabular}{cc}
    \toprule
    \textbf{Hyperparameter} & \textbf{Value} \\
    \midrule
    learning rate & $2\!\cdot\!10^{-4}$ \\
    weight decay & 0.1 \\
    batch size & 512 \\
    training steps & $10^6$ \\
    warmup steps & $10^4$ \\
    number of layers & $20$ \\
    embedding dimension & $512$ \\
    dropout & 0.1 \\
    RWSE dimension & 20 \\
    \bottomrule
    \end{tabular}

\end{table}

\subsection{Multi-task pre-training with encoder-processor-decoder}\label{app:gfm}
Here, we outline details for our experiments with the encoder-processor-decoder setup.

\subsubsection{Model architectures}
We consider an encoder-processor-decoder setup and two different processors, an MPNN with GINE~\citep{hu2020pretraining} layers and a graph transformer derived from Graphormer~\citep{ying2021transformers}. As a common practice in the case of transformer architectures on graphs, we also experiment with injecting node-wise structural encodings, particularly RWSEs~\citep{dwivedigraph}. In what follows, we detail the encoder-processor-decoder setup.

\paragraph{Encoder}
Using task-specific encoders, we embed node and edge features into a common embedding dimension $d \in \mathbb{N}^+$ for both architectures. If no node or edge features are available, we use learnable vectors that we train jointly with the architecture. Following standard practice in (graph) transformer encoders \citep{ying2021transformers}, we add a \texttt{[cls]} token from which we read out graph-level representations.

\paragraph{Processor}
Subsequently, a processor network computes node- and graph-level representations from the embedded node, edge features, and graph structure. 
Given a graph $G$, both the MPNN and graph transformer update node representations $\mathbf{X} \in \mathbb{R}^{n \times d}$ at each layer as
\begin{align*}
    \mathbf{X}' &\leftarrow \mathbf{X} + \phi(\mathsf{LayerNorm}(\mathbf{X}), G), \\
    \mathbf{X}''  &\leftarrow \mathbf{X}' + \mathsf{MLP}(\mathsf{LayerNorm}(\mathbf{X}')),
\end{align*}
where $\textsf{MLP}$ is a two-layer MLP with GELU non-linearity \citep{hendrycks2016gelu} and $\phi(\cdot, G)$ is either a graph convolution or attention, conditioned on $G$. Graph convolution is implicitly conditioned via message-passing over the local neighborhood. 

In the case of attention, we add a graph-aware attention bias to the unnormalized attention matrix. Concretely, the graph transformer layer computes full multi-head scaled-dot-product attention over node-level tokens with a soft-attention bias computed from the edge features of the graph. The attention bias is a tensor $\mathbf{B} \in \mathbb{R}^{L \times L \times h}$, where $L \in \mathbb{N}^+$ is the number of tokens and $h \in \mathbb{N}^+$ is the number of attention heads. In particular, we compute a separate attention bias for each attention head.
For a graph with $n$ nodes, we set $L \coloneqq n + 1$ (accounting for the \texttt{[cls]} tokens). For simplicity, we write $i \in \mathbb{N}^+$ to indicate the $i$th node in an arbitrary but fixed node ordering. We refer to the \texttt{[cls]} tokens as node $n+1$. Further, only for the graph transformer, we use a maximum context size of 8192 and remove additional nodes that exceed this size.
We then compute the attention bias $\mathbf{B}$ such that for all edges $(i, j)$,
\begin{equation*}
    \mathbf{B}_{ij} \coloneqq \mathbf{W} \cdot \mathbf{e}_{ij},
\end{equation*}
where $\mathbf{e}_{ij} \in \mathbb{R}^{d}$ is the edge feature of $(i, j)$ and $\mathbf{W} \in \mathbb{R}^{d \times h}$ is a learnable weight matrix. Again, we omit bias terms for clarity. If no edge exists between nodes $i$ and $j$, we set $\mathbf{B}_{ij}$ to all-zeros.
For the \texttt{[cls]} token, we use learnable vectors $\mathbf{e}_\text{in}, \mathbf{e}_\text{out} \in \mathbb{R}^d$ as the attention bias for in- and out-coming edges, respectively, i.e., we set
\begin{align*}
    \mathbf{B}_{(n+1)j} &\coloneqq \mathbf{e}_\text{in}, \\
    \mathbf{B}_{i(n+1)} &\coloneqq \mathbf{e}_\text{out}.
\end{align*}
Finally, we add $\mathbf{B}$ as a soft bias to the unnormalized attention matrix, that is, before applying softmax.

\paragraph{Decoder}
Lastly, we apply a decoder network that makes task-specific predictions. In our experiments, we used the same MLP layout for all decoders. In particular, given a representation vector $\mathbf{x} \in \mathbb{R}^d$, we define our decoder MLP as
\begin{equation*}
    \mathbf{W}_2\mathsf{LayerNorm}(\mathsf{GELU}(\mathbf{W}_1\mathbf{x})),    
\end{equation*}
where $\mathbf{W}_1 \in \mathbb{R}^{d \times d}$, $\mathbf{W}_2 \in \mathbb{R}^{d \times o}$ are learnable weight matrices, with $o \in \mathbb{N}^+$ the task-specific output dimensions (e.g., the number of classes in a classification task) and a $\mathsf{GELU}$ non-linearity \citep{hendrycks2016gelu}. We omit bias terms for clarity.

\subsubsection{Multi-task pre-training}

Here, we outline details on the multi-task pre-training.

\paragraph{Training loop and optimization parameters} We perform multi-task pre-training by using data loaders for all tasks and accumulating gradients from each task at each iteration of the training loop, effectively simulating a \say{heterogeneous batch} of data from all available tasks. We train on bfloat16 with clipped gradients and a cosine learning rate scheduler.
\begin{table}[h!]
  \caption{Datasets in the pretraining mix.}
    \label{tab:pretrain-mix}
    \centering
    \resizebox{\linewidth}{!}{
    \begin{tabular}{lcccc}
        \toprule
        \textbf{Dataset} & \textbf{Domain} & \textbf{Task} & \textbf{Avg. \# nodes} & \textbf{Avg. \# edges} \\
        \midrule
        \textsc{COCO-SP} & Vision (Super Pixels) & Semantic Segmentation & 476.88 & 2\,693.67 \\
        \textsc{MalNetTiny} & Cybersecurity (Function Calls Graphs) & Malware Detection & 1\,410.3 & 2\,859.9 \\
        \textsc{PCQM4Mv2} & Chemistry (Small 2D Molecules) & HOMO-LUMO Gap Prediction & 14.1 & 14.6 \\
        \bottomrule
    \end{tabular}
    }

\end{table}

\paragraph{Pretraining mix} As already mentioned in the main text, the pretraining mix is formed by the datasets described in~\Cref{tab:pretrain-mix}, which differ in domain, task, and structural properties.

\begin{itemize}
    \item \textsc{COCO-SP}~\cite{Dwi+2022} is a dataset of sparse, medium-sized graphs encoding images at the super-pixel level. Nodes, i.e., super-pixels, are attributed with pixel value statistics and center-of-mass coordinates. The task is to predict, for each super-pixel, a semantic segmentation label.
    \item \textsc{PCQM4Mv2}~\cite{hu2ogb} comprises many small molecular graphs for which the task is to predict the HOMO-LUMO energy gap. Interestingly, only accessing 2D molecular information is practically relevant in this setting, as calculating 3D structures requires expensive DFT-based geometry optimization.\footnote{See \url{https://ogb.stanford.edu/docs/lsc/pcqm4mv2/}.}
    \item \textsc{MalNet-Tiny}\citep{freitas2021malnet} includes a relatively small number of larger graphs encoding function calls, with the task being to predict their association with malicious code execution. These graphs are entirely unattributed.
\end{itemize}

\paragraph{Hyperparameter tuning} We sweep the learning rate over $\{ 4\cdot 10^{-5}, 7\cdot 10^{-5}, \ldots, 1\cdot 10^{-3} \}$ for graph transformers and $\{ 4\cdot 10^{-5}, 7\cdot 10^{-5}, \ldots, 1\cdot 10^{-2} \}$ for GINE and train for 100K gradient steps. We pick the pre-trained checkpoint based on the best overall validation loss, which we compute as the sum of all three task losses.

\subsubsection{Single-task fine-tuning}

Here, we outline details on the single-task fine-tuning.

\paragraph{Architectural details}  Across all finetuning experiments, the prediction heads (i.e., the decoders) are initialized and trained from scratch, while the (pre-trained) processors are kept frozen. If a downstream task shares the same node and/or edge features with a pretraining dataset, we reuse the corresponding (pre-trained) encoders, which are also frozen during finetuning. Otherwise, a new encoder is initialized and trained from scratch. Note that downstream datasets with featureless nodes share identical (pre-trained) encoders; see, e.g., \textsc{Stargazers} below.

In all cases, we run a standard single-task finetuning on bfloat16 with clipped gradients and a cosine learning rate scheduler. 

\begin{table}[h!]
    \centering
      \caption{Datasets considered for downstream applications.}
    \label{tab:finetune-datasets}
    \begin{tabular}{lcccc}
        \toprule
        \textbf{Dataset} & \textbf{Domain} & \textbf{Task} & \textbf{Avg. \# nodes} & \textbf{Avg. \# edges}\\
        \midrule
        \textsc{PascalVOC-SP} & Vision (Super Pixels) & Semantic Segmentation & 479.40 & 2\,710.5 \\
        \textsc{Peptides-struct} & Chemistry (Peptides) & 3D-Structure Property Prediction & 150.9 & 307.3 \\
        %\textsc{Peptides-func} & Chemistry (Peptides) & Functional Prediction & 150.9 & 307.3 \\
        \textsc{Stargazers} & Github Communities & Social Network Classification & 113.79 & 234.64 \\
        %ogbg-code2 ? & Code (Abstract Syntax Trees) & Method Name Subtoken Prediction & 125.2 & ??? \\
        \bottomrule
    \end{tabular}
  
\end{table}

\paragraph{Downstream (finetuning) datasets} The datasets considered as downstream applications for our finetuning experiments are enlisted and concisely described in~\Cref{tab:finetune-datasets}. Again, they vary widely in domains, tasks, and structural features while encompassing various levels of similarity to the datasets in the pretraining mix. In particular:
\begin{itemize}
    \item \textsc{PascalVOC-SP}~\cite{Dwi+2022} is aligned with \textsc{COCO-SP} in most aspects: domain, task and structure. Here, we can reuse the pre-trained encoder of \textsc{COCO-SP}. 
    \item \textsc{Peptides-struct}~\cite{Dwi+2022} comprise molecular graphs and belong to the same broad chemical domain of \textsc{PCQM4Mv2}, from which our downstream model reuses the feature encoder. However, these molecular graphs are distinct (they represent chains of amino acids) and structure (they are larger and more elongated, with higher diameter values). The task is also different in that it pertains to the prediction of 3D structural features rather than quantum properties.
    \item \textsc{Stargazers} \citep{rozemberczki2020karateclub, Mor+2020} comprises social networks formed by GitHub developers who have starred at least 10 repositories, connected by `following' relations. The task is to classify these social networks as belonging to either Web or Machine Learning developers. This dataset is completely different in task and domain from any of the datasets considered in the pretraining mix. No node or edge features are available, so our model reuses the same node encoder pre-trained on the featureless MalNetTiny.
\end{itemize}

\paragraph{Training setup} In this setting, we are interested in measuring the sample efficiency of our models, aiming to study if and when pre-training is beneficial. Accordingly, we train for 1K, 3K, and 10K steps, setting the batch size to correspond to roughly 2, 10, and 30 epochs. The same setting is employed for reference single-task baselines, trained from scratch on the same amount of data.

\paragraph{Hyperparameter tuning} The learning rate is the only tuned hyper-parameter; we sweep over 3 orders of magnitude in $\{1\cdot10^{-5}, 4\cdot10^{-5}, 7\cdot10^{-5}, \ldots, 1\cdot10^{-2}\}$ for each fine-tuning regime. The single-task baselines are always sized in a way to total the same number of parameters of their multi-task pre-trained counterparts.

\begin{figure*}[!t]
    \centering    \includegraphics[width=\linewidth]{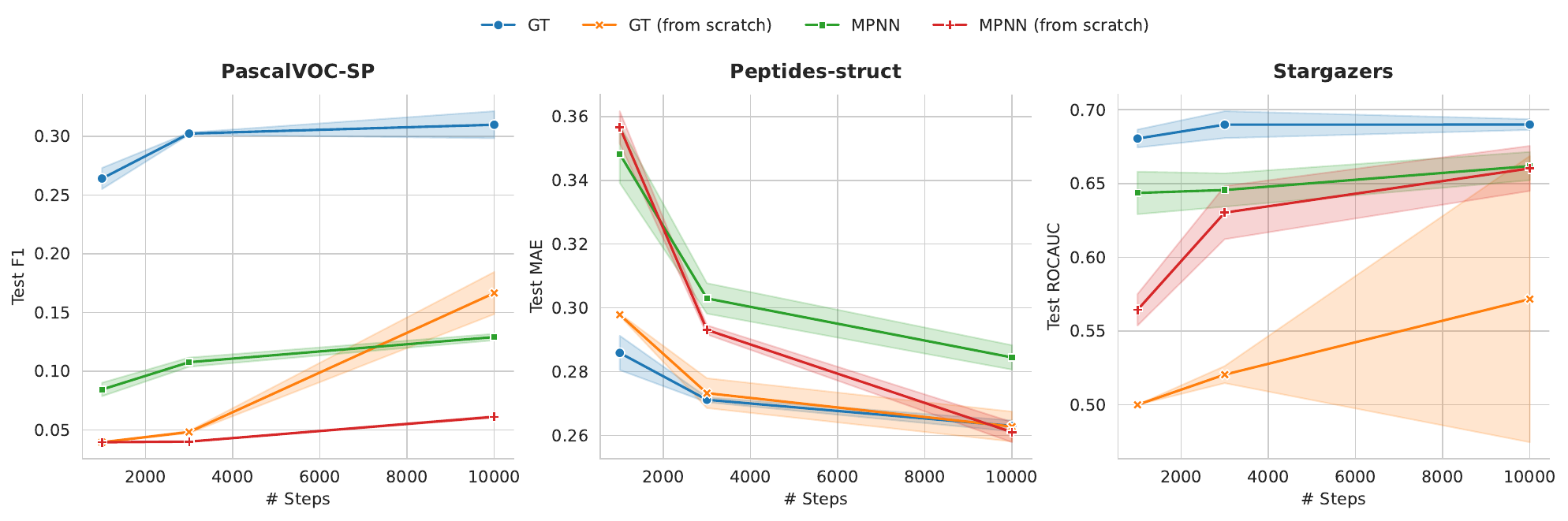}
    \caption{\textbf{Highlights} from our fine-tuning results on three downstream datasets with varying numbers of fine-tuning steps and varying degrees of similarity to the upstream mix.}
    \label{fig:fm_downstream}
\end{figure*}

\begin{figure}
    \centering
    \includegraphics[width=\linewidth]{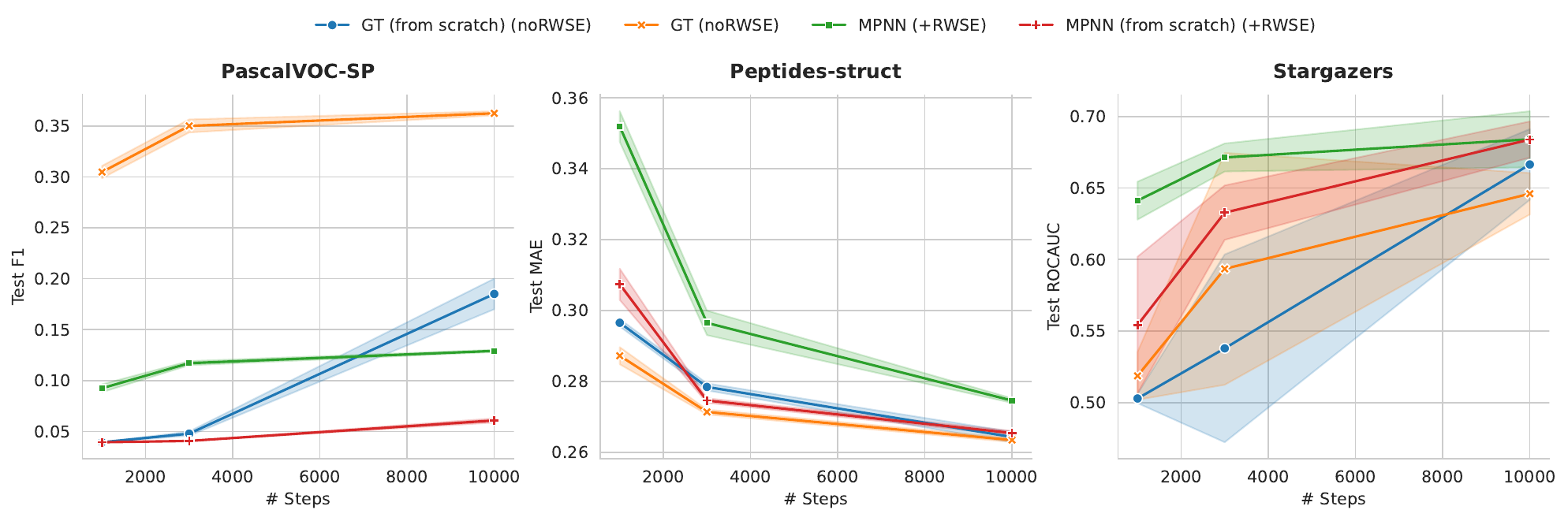}
    \caption{Additional downstream results for MPNN+RWSE and graph transformers without additional structural encodings. Note that the GT is still graph-aware due to the soft attention bias.}
    \label{fig:fm_downstream_extra}
\end{figure}

\paragraph{Additional results} In addition to \Cref{fig:fm_downstream}, we provide additional fine-tuning results in \Cref{fig:fm_downstream_extra}, where we compare the GNN with additional RWSE, as well as the graph transformer without additional structural encodings.

\subsection{Variance of results reported on \textsc{\textsc{Enzymes}}}
\label{subsec:enyzmes}

\begin{figure}[t]
  \centering
  \includegraphics[width=.7\textwidth]{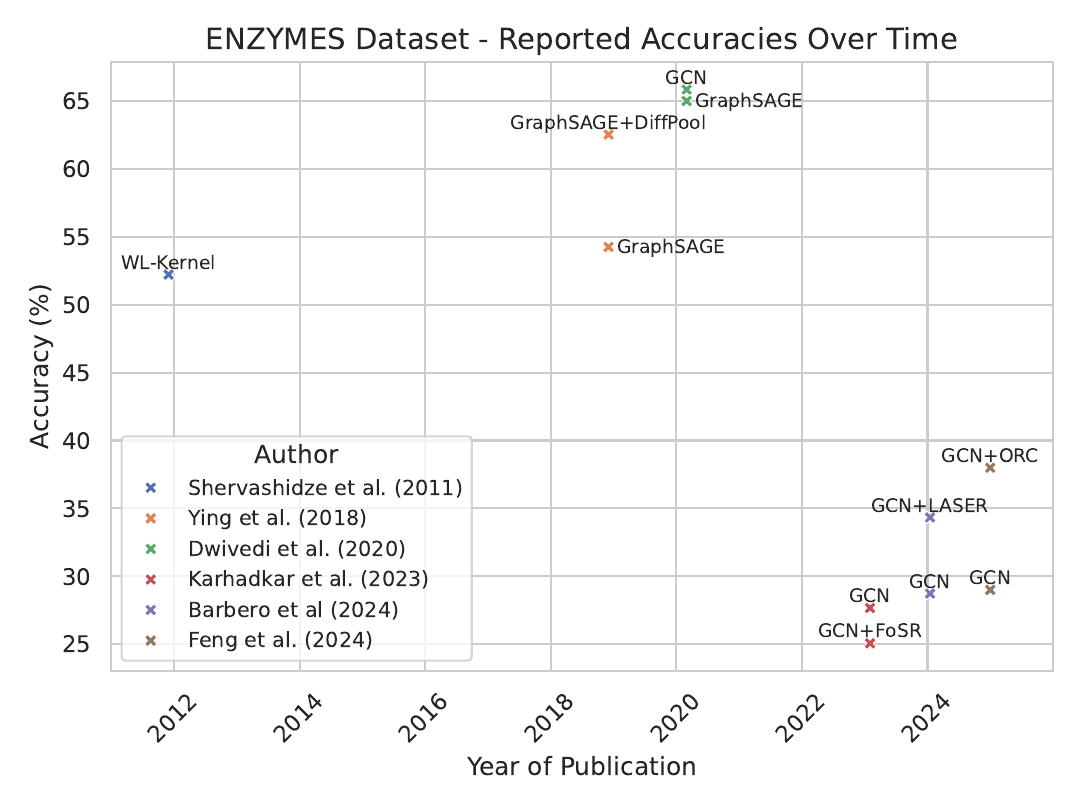}
  \caption{Test accuracy reported on the \textsc{\textsc{Enzymes}} dataset over the past twelve years in various publications: \citet{She+2011, ying2018hierarchical,Dwi+2020,karhadkar2023fosr,barbero2024localityaware,feng2024graph}}
  \label{fig:Enzymes}
\end{figure}

In \Cref{sec:culture}, we discussed problematic practices prevalent in experimental evaluations of GNNs. A common problem is using small, high-variance datasets without an established evaluation protocol. For some datasets, the numbers reported throughout the literature vary substantially, resulting in an inconsistent and confusing representation of model performance.  Here, we illustrate this problem for the commonly used \textsc{Enzymes} dataset \citep{Mor+2020} as an example of how extreme reported performance measurements vary.

In \Cref{fig:Enzymes}, we plot the reported test accuracy of different graph learning publications on the \textsc{Enzymes} dataset against the year of publication.
We include results from various lines of work, such as graph kernels \cite{She+2011}, GNN benchmarking \cite{Dwi+2020}, graph pooling  \cite{ying2018hierarchical, feng2024graph}, and graph rewiring \cite{karhadkar2023fosr, barbero2024localityaware}.
The results reveal an interesting trend.
Older kernel-based methods, such as the WL kernel, achieved a baseline accuracy of around 52\%.
Initial evaluations of MPNN models outperformed this baseline, often achieving over 60\%, even with simple GCN-based models. However, newer publications from 2023 and 2024 often fall short of these results by a significant margin, sometimes reporting less than 30\% classification accuracy, even when evaluating similar GCN-based architectures.
In other words, the results reported for the same base architecture can vary by a factor of two across publications.

There are several causes for this extreme variance. 
First of all, there is no consistent evaluation setup.
While older publications typically used stratified 10-fold cross-validation as an evaluation protocol, newer results are often based on repeated random 80/10/10 splits, which are prone to be more noisy. This difference explains the performance variance to some degree but does not account for the sharp drop in the reported accuracy of more recent publications. Instead, many recent works seem to run experiments with suboptimal hyper-parameter choices, resulting in a significant loss in performance for compared models. For example, \citet{barbero2024localityaware} configure the training to only last 100 epochs, which is too short to allow for model convergence on a dataset as small as \textsc{Enzymes}.

While each publication is internally consistent in the sense that it applies the same experimental setup to the methods it compares, one can argue that this is not sufficient when measurements vary this drastically over time.
The lack of standardization risks conflating methodological improvements with artifacts of experimental design. Ensuring cross-study consistency---through adherence to shared protocols and rigorous benchmarking on more suitable datasets---is critical to fostering trust in reported results and enabling clear advancements in graph learning research.

\end{document}